\documentclass[10pt, a4paper, fleqn]{article}
\usepackage{lrec2006}
\usepackage{graphicx}
\usepackage{url}
\expandafter\def\expandafter\UrlBreaks\expandafter{\UrlBreaks
  \do\a\do\b\do\c\do\d\do\e\do\f\do\g\do\h\do\i\do\j%
  \do\k\do\l\do\m\do\n\do\o\do\p\do\q\do\r\do\s\do\t%
  \do\u\do\v\do\w\do\x\do\y\do\z\do\A\do\B\do\C\do\D%
  \do\E\do\F\do\G\do\H\do\I\do\J\do\K\do\L\do\M\do\N%
  \do\O\do\P\do\Q\do\R\do\S\do\T\do\U\do\V\do\W\do\X%
  \do\Y\do\Z}

\usepackage{hyperref}
\usepackage{amsmath}

\usepackage{array}
\newcolumntype{P}[1]{>{\centering\arraybackslash}p{#1}}

\title{To What Extent are Name Variants Used as Named Entities in Turkish Tweets?}

\name{Dilek K\"u\c{c}\"uk}

\address{Electrical Power Technologies Department\\
T\"UB\.ITAK Energy Institute\\
Ankara{--}Turkey\\
dilek.kucuk@tubitak.gov.tr\\}

\abstract{Social media texts differ from regular texts in various aspects. One of the main differences is the common use of informal name variants instead of well-formed named entities in social media compared to regular texts. These name variants may come in the form of abbreviations, nicknames, contractions, and hypocoristic uses, in addition to names distorted due to capitalization and writing errors. In this paper, we present an analysis of the named entities in a publicly-available tweet dataset in Turkish with respect to their being name variants belonging to different categories. We also provide finer-grained annotations of the named entities as well-formed names and different categories of name variants, where these annotations are made publicly-available. The analysis presented and the accompanying annotations will contribute to related research on the treatment of named entities in social media.\\ \newline }

\begin{document}

\maketitleabstract

\section{Introduction}\label{sec:intro}

Automatic extraction and classification of named entities in natural language texts (i.e., named entity recognition (NER)) is a significant topic of natural language processing (NLP), both as a stand-alone research problem and as a subproblem to facilitate solutions of other related NLP problems. NER has been studied for a long time and in different domains, and there are several survey papers on NER including \cite{marrero2013named}.\\

Conducting NLP research (such as NER) on microblog texts like tweets poses further challenges, due to the particular nature of this text genre. Contractions, writing\slash grammatical errors, and deliberate distortions of words are common in this informal text genre which is produced with character limitations and published without a formal review process before publication. There are several studies that propose tweet normalization schemes \cite{han2011lexical} to alleviate the negative effects of such language use in microblogs, for the other NLP tasks to be performed on the normalized microblogs thereafter. Yet, particularly regarding Turkish content, a related study on NER on Turkish tweets \cite{kucuk2014experiments} claims that normalization before the actual NER procedure on tweets may not guarantee improved NER performance.\\

Identification of name variants is an important research issue that can help facilitate tasks including named entity linking \cite{weichselbraun2019name} and NER, among others. Name variants can appear due to several reasons including the use of abbreviations, contracted forms, nicknames, hypocorism, and capitalization\slash writing errors \cite{weichselbraun2019name}. The identification and disambiguation of name variants have been studied in studies such as \cite{driscoll2007disambiguation} and \cite{weichselbraun2019name}, where resource-based and\slash or algorithmic solutions are proposed.\\

In this paper, we consider name variants from the perspective of a NER application and analyze an existing named entity-annotated tweet dataset in Turkish described in \cite{kucuk2014named}, in order to further annotate the included named entities with respect to a proprietary name variant categorization. The original dataset includes named annotations for eight types: \texttt{PERSON}, \texttt{LOCATION}, \texttt{ORGANIZATION}, \texttt{DATE}, \texttt{TIME}, \texttt{MONEY}, \texttt{PERCENT}, and \texttt{MISC} \cite{kucuk2014named}. However, in this study, we target only at the first three categories which amounts to a total of 980 annotations in 670 tweets in Turkish. We further annotate these 980 names with respect to a name variant categorization that we propose and try to present a rough estimate of the extent at which different named entity variants are used as named entities in Turkish tweets. The resulting annotations of named entities as different name variants are also made publicly available for research purposes. We believe that both the analysis described in the paper and the publicly-shared annotations (i.e., a tweet dataset annotated for name variants) will help improve research on NER, name disambiguation, and name linking on Turkish social media posts.\\

The rest of the paper is organized as follows: In Section 2, an analysis of the named entities in the publicly-available Turkish tweet dataset with respect to their being name variants or not is presented together with the descriptions of name variant categories. In Section 3, details and samples of the related finer-grained annotations of named entities are described and Section 4 concludes the paper with a summary of main points.

\section{An Analysis of Turkish Tweets for Name Variants Included}\label{sec:analysis}

Although NER is an NLP topic that has been studied for a long time, currently, the target genre of the related studies has shifted from well-formed texts such as news articles to microblog texts like tweets \cite{ritter2011named}. Following this scheme (mostly) on English content, NER research on other languages like Turkish has also started to target at tweets \cite{kucuk2014named,kucuk2014experiments}. A named entity-annotated dataset consisting of Turkish tweets is described in \cite{kucuk2014named} and the results of NER experiments on Turkish tweets are presented in \cite{kucuk2014experiments}. Interested readers are referred to \cite{kucuk2017namedsurvey} which presents a survey of named entity recognition on Turkish, including related work on tweets.\\

In this study, we analyze the basic named entities (of type \texttt{PERSON}, \texttt{LOCATION}, and \texttt{ORGANIZATION}, henceforth, PLOs) in the annotated dataset compiled in \cite{kucuk2014named}, with respect to their being well-formed canonical names or name variants. The dataset includes a total of 1.322 named entity annotations, however, 980 of them are PLOs (457 \texttt{PERSON}, 282 \texttt{LOCATION}, and 241 \texttt{ORGANIZATION} names) and are the main focus of this paper. These 980 PLOs were annotated within a total of 670 tweets.\\

We have extracted these PLO annotations from the dataset and further annotated them as belonging to one of the following eight name variant categories that we propose. We should note that a particular name can belong to several categories and therefore, there may be multiple category labels assigned to it. However, the number of category labels does not exceed two in our case, i.e., each name is annotated with either one or two labels in the resulting dataset.

\begin{itemize}
    \item \emph{WELL-FORMED}: This category comprises those names which are written in their open and canonical form without any distortions, conforming to the capitalization and other writing rules of Turkish. In Turkish, each of the tokens of names are written with their initial letters capitalized. However, those names written all in uppercase are also considered within this category as they cannot be considered as writing errors.
    \item \emph{ABBREVIATION}: This category represents those names which are provided as abbreviations. This usually applies to named entities of \texttt{ORGANIZATION} type. But, these abbreviations can include writing errors due capitalization or characters with diacritics, as will be explained below. Hence, those names annotated as \emph{ABBREVIATION} can also have an additional category label as \emph{CAPITALIZATION} or \emph{DIACRITICS}.
    \item \emph{CAPITALIZATION}: This category includes those names distorted due to not conforming to the capitalization rules of Turkish. As pointed out above, initial letters of each of the tokens of a named entity are capitalized in Turkish. Additionally, abbreviations of names are generally all in uppercase. Those names not conforming to these rules are marked with the \emph{CAPITALIZATION} label, denoting a capitalization issue.
    \item \emph{DIACRITICS}: There are six letters with diacritics in Turkish alphabet \{\c{c}, \u{g}, {\i}, \"{o}, \c{s}, \"{u}\} which are sometimes replaced with their counterparts without diacritics \{c, g, i, o, s, u\}, in informal texts like microblogs \cite{kucuk2014experiments}. Very rarely, the opposite (and perhaps unintentional) replacements can be observed again in informal texts (this time at least one character without diacritics is replaced with a character having diacritics in a word). Named entities including such writing errors are assigned the category label of \emph{DIACRITICS}.
    \item \emph{HASHTAG-LIKE}: Another name variant type is the case where the whitespaces in the names are removed, so they appear like hashtags, and sometimes they are actually hashtags. Such phenomena are annotated with the category label of \emph{HASHTAG-LIKE}.
    \item \emph{CONTRACTED}: This category represents those name variants in which the original name is contracted, by leaving out some of its tokens. Since users like to produce and publish instantly on social media, they tend to contract especially those long organization names, mostly by using its initial token only. Such name variants are annotated as \emph{CONTRACTED}.
    \item \emph{HYPOCORISM}: Hypocorism or hypocoristic use is defined as the phenomenon of deliberately modifying a name, in the forms of nicknames, diminutives, and terms of endearment, to show familiarity and affection \cite{newman1992hypocoristic,driscoll2013computational}. An example hypocoristic use in English is using \emph{Bobby} instead of the name \emph{Bob} \cite{newman1992hypocoristic}. Such name variants observed in the tweet dataset are marked with the category label of \emph{HYPOCORISM}.
    \item \emph{ERROR}: This category denotes those name variants which have some forms of writing errors, excluding issues related to capitalization, diacritics, hypocorism, and removing whitespaces to make names appear like hashtags. Hence, names conforming to this category are labelled with \emph{ERROR}.
\end{itemize}

The following subsection includes examples of the above name variant categories in the Turkish tweet dataset analyzed, in addition to statistical information indicating the share of each category in the overall dataset.

\section{Finer-Grained Annotation of Named Entities}\label{sec:dataset}

We have annotated the PLOs in the tweet dataset (already-annotated for named entities as described in \cite{kucuk2014named}) with the name variant category labels of \emph{WELL-FORMED}, \emph{ABBREVIATION}, \emph{CAPITALIZATION}, \emph{DIACRITICS}, \emph{HASHTAG-LIKE}, \emph{CONTRACTED}, \emph{HYPOCORISM}, and \emph{ERROR}, as described in the previous subsection. Although there are 980 PLOs in the dataset, since 44 names have two name variant category labels, the total number of name variant annotations is 1,024.\\

\begin{figure*}[h!]
\center \scalebox{0.31}
{\fbox{\includegraphics{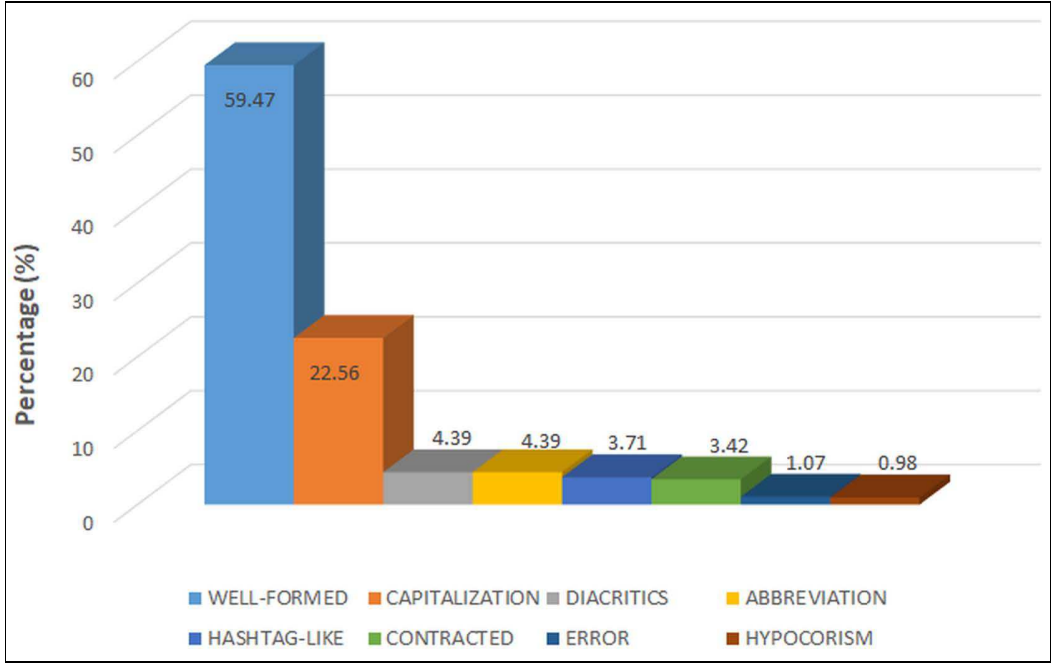}}}\caption{Statistical Information for Each Named Entity Variant Category in the Turkish Tweet Dataset.}\label{fig:graph}
\end{figure*}

The percentages of the category labels in the final annotation file are provided as a bar graph in Figure \ref{fig:graph}. As indicated in the figure, about 60\% of all named entities are well-formed and hence about 40\% of them are not in their canonical open form or do not conform to the capitalization\slash writing errors regarding named entities in Turkish.\\

The most common issue is the lack of proper capitalization of names in tweets, revealed with a percentage of 22.56\% names annotated with the \emph{CAPITALIZATION} label. For instance, people write \emph{istanbul} instead of the correct form \emph{\.{I}stanbul} and \emph{ankara} instead of \emph{Ankara} in their tweets.\\

The number of names having issues about characters with diacritics is 45, and similarly there are 45 abbreviations (of mostly organization names) in the dataset. As examples of names having issues with diacritics, people use \emph{Kutahya} istead of the correct form \emph{K\"{u}tahya}, and similarly \emph{Besiktas} instead of \emph{Be\c{s}ikta\c{s}}. Abbreviations in the dataset include national corporations like \emph{TRT} and \emph{SGK}, and international organizations like \emph{UEFA}.\\

Instances of the categories of \emph{HASHTAG-LIKE} and \emph{CONTRACTED} are observed in 38 and 35 names, respectively. A sample name variant marked with \emph{HASHTAG-LIKE} is \emph{SabriSar{\i}o\u{g}lu} where this person name should have been written as \emph{Sabri Sar{\i}o\u{g}lu}. A contracted name instance in the dataset is \emph{Diyanet} which is an organization name with the correct open form of \emph{Diyanet \.{I}\c{s}leri Ba\c{s}kanl{\i}\u{g}{\i}}.\\

The instances of \emph{HYPOCORISM} and \emph{ERROR} are comparatively low, where 10 instances of hyprocorism and 11 instances of other errors are seen in the dataset. An instance of the former category is \emph{Nazl{\i}\c{s}} which is a hypocoristic use of the female person name \emph{Nazl{\i}}. An instance of the \emph{ERROR} category is the use of \emph{FENEBAH\c{C}E} instead of the correct sports club name \emph{FENERBAH\c{C}E}.\\

Overall, this finer-granularity analysis of named entities as name variants in a common Turkish tweet dataset is significant due to the following reasons.
\begin{itemize}
  \item The analysis leads to a breakdown of different named entity variants into eight categories. Although about 60\% of the names are in their correct and canonical forms, about 40\% of them either appear as abbreviations or suffer from a deviation from the standard form due to multiple reasons including violations of the writing rules of the language. Hence, it provides an insight about the extent of the use of different name variants as named entities in Turkish tweets.
  \item The use of different name variants is significant for several NLP tasks including NER on social media, name disambiguation and linking. A recent and popular research topic that may benefit from patterns governing name variants is stance detection, where the position of a post owner towards a target is explored, mostly using the content of the post \cite{mohammad2016semeval}. A recent study reports that named entities can be used as improving features for the stance detection task \cite{kucuk2017joint_stance_ner}. Hence, an analysis of name variants can contribute to the algorithmic\slash learning-based proposals for these research problems.
\end{itemize}

The name variant annotations described in the study are made publicly available at \url{https://github.com/dkucuk/Name-Variants-Turkish-Tweets} as a text file, for research purposes. Each line in the annotation file denotes triplets, separated by semicolons. The first item in each triplet is the tweet id, the second item is another triplet denoting the already-existing named entity boundaries and type, and the final item is a comma-separated list of name variant annotations for the named entity under consideration. Below provided are two sample lines from the annotation file. The first line indicates a person name (between the non-white-space characters of 0 and 11 in the tweet text) annotated with CAPITALIZATION category, as it lacks proper capitalization. The second line denotes an organization name (between the non-white-space characters of 0 and 19 in the tweet) which has issues related to characters with diacritics and proper capitalization.\\

\texttt{360731728177922048;0,11,PERSON;CAPITALIZATION}\\
\texttt{360733236961349636;0,19,ORGANIZATION;DIACRITICS,CAPITALIZATION}

\section{Conclusion}\label{sec:conc}

This paper focuses on named entity variants in Turkish tweets and presents the related analysis results on a common named-entity annotated tweet dataset in Turkish. The named entities of type person, location, and organization names are further categorized into eight proprietary name variant classes and the resulting annotations are made publicly available. The results indicate that about 40\% of the considered names deviate from their standard canonical forms in these tweets and the categorizations for these cases can be used by researchers to devise solutions for related NLP problems. These problems include named entity recognition, name disambiguation and linking, and more recently, stance detection.

\end{document}